\documentclass[conference]{IEEEtran}
\ifCLASSINFOpdf
\else
\fi

\usepackage{graphicx}
\usepackage{url}
\usepackage{bm}
\usepackage{subfigure}
\usepackage{algpseudocode}
\usepackage{multirow}
\usepackage{url}
\usepackage{balance}
\usepackage{epstopdf}

\usepackage{cite}
\usepackage{amsmath,amssymb,amsfonts}
\usepackage{algorithm}
\usepackage{algorithmicx}
\usepackage{textcomp}
\usepackage{xcolor}
\def\BibTeX{{\rm B\kern-.05em{\sc i\kern-.025em b}\kern-.08em
    T\kern-.1667em\lower.7ex\hbox{E}\kern-.125emX}}
\begin{document}
%


\title{Efficient delivery of Robotics Programming educational content using Cloud Robotics}


\author{\IEEEauthorblockN{Sean Murphy,
Leonardo Militano,
Giovanni Toffetti and
Remo Maurer}
\IEEEauthorblockA{\textit{Zurich University of Applied Sciences}\\
Switzerland\\
Email:  $[ murp \mid milt \mid toff \mid murm ]$ @zhaw.ch}}

\maketitle

\begin{abstract}
In this paper, we report on our use of cloud-robotics solutions to teach a Robotics Applications Programming course at Zurich University of Applied Sciences (ZHAW).
The usage of Kubernetes based cloud computing environment combined with real robots -- turtlebots and Niryo arms -- allowed 
us to: 1) minimize the set up times required to provide a Robotic Operating System (ROS) simulation and development environment to all students 
independently of their laptop architecture and OS; 2) provide a seamless ``simulation to real'' experience preserving the 
exciting experience of writing software interacting with the physical world; and 3) sharing GPUs across multiple 
student groups, thus using resources efficiently.

We describe our requirements, solution design, experience working with the solution in the educational
context and areas where it can be further improved. This may be of interest to other educators who may want 
to replicate our experience.

\end{abstract}

\begin{IEEEkeywords}
 Cloud Robotics, Kubernetes, Robotic Applications, Edge Computing.
\end{IEEEkeywords}

%
\IEEEpeerreviewmaketitle

\section{Introduction}
\label{sec_1}


The School of Engineering at ZHAW has been offering the ``Robotic Applications Programming'' (RAP) course to bachelor students since 2021. The course is intended for IT Bachelor students as a way to 1) learn how to program applications using ROS-based robots, and 2) leverage interdisciplinary knowledge acquired during the study programme (e.g., Artificial Intelligence, computer vision, distributed systems, cloud, Operating Systems, web- and mobile-development) and integrate them to achieve autonomous robotic behavior. We focus on ROS as it is currently the most used framework, it is Open Source, and has ever-increasing capabilities with many contributions of advanced algorithm implementations and robotic simulation packages \cite{toff}, \cite{garcia}.

The course is organized in three main sections as follows: (1) students first learn ROS fundamentals (communication primitives and building ROS packages) and robotics (e.g., basic robotic Hardware, robot models and visualization, coordinate frames and transformations, controllers); (2) additional base capabilities are learned (e.g., Simultaneous Localization and Mapping - SLAM, navigation, perception, arm motion planning and control); and (3) finally these are combined to build a practical application for the yearly challenge. We do not explicitly consider mechanical engineering aspects nor system integration aspects that complete the robotics engineering field.
This year's challenge is inspired by the DARPA Subterranean Challenge\footnote{https://www.subtchallenge.com/}: students will have to write software to control an autonomous mobile manipulator -- a simulated Summit XL\footnote{https://robotnik.eu/products/mobile-robots/summit-xl-en-2/} with a UR-5 arm\footnote{https://www.universal-robots.com/products/ur5-robot/} -- in an unknown environment performing mapping, pose estimation and collection of known objects and returning all objects to the starting location.

During the course of the semester, students apply the theoretical concepts they learn in class to lab sessions; the earlier lab sessions use simulated robots for quick software development cycles and the later lab sessions run the same software to control real robots - 6 turtlebot3's are used for SLAM/navigation and 3 Niryo arms are used for grasping.

In order for the students to concentrate on course content and minimize the time they would need for system set up and configuration, we needed to prepare some teaching infrastructure. We had the following key requirements:
\begin{itemize}
\item{\textbf{R1:}} Provide a consistent collaborative environment for group work across multiple access devices (tablets, laptops with different Operating Systems and CPU architectures);
\item{\textbf{R2:}} Support Simulation with a realistic simulated-to-real time ratio and frame-rate;
\item{\textbf{R3:}} Support transitioning from the simulated environment to real world robots with minimal effort
\end{itemize}

The main contribution of this paper then is the system design which meets these requirements. Possible technologies are described, evaluated and the design choices for the final solution are discussed.  

The paper is structured as follows. In section \ref{sec:related_work}, we review related work
broadly classifying solutions into simulation focused solutions, hardware focused solutions and hybrid solutions. Section \ref{sec:solution_design} describes our solution
including the basic components and how they fit together. In section \ref{sec:demo_eval} we discuss our experience using the platform in the classroom environment.
Section \ref{sec:issues_challenges} discusses open issues with the current solution and finally there is a conclusion in section \ref{sec:conclusion}.

\section{Related Work}
\label{sec:related_work}
The interest in robotics engineering has been growing rapidly over the last few years. For hobbyists, students and professionals, the amount of robotics practitioners has steadily grown and with it the available educational content. At the same time, educational institutions at all levels are working hard to adapt their instructional programs and learning paths to integrate robotic technologies. For instance, Educational Robotics (ER) is a modern teaching practice that the teacher engages in, using robots as a tool for designing and integrating the educational process. ER was identified as an educational resource through which students acquire knowledge of different disciplines and improve their attitude and interest in STEAM disciplines (Science, Technology, Engineering, Arts and Mathematics) \cite{schina}, \cite{curto}. 

 Depending on the educational level and the requirements for professional knowledge of robotic application development, different teaching and learning approaches can be identified. Here, we categorize them as follows: i) simulation-based; ii) hardware-based; and iii) combination of simulation- and hardware-based solutions.

\textbf{Simulation-based learning} leverage software tools and programming languages to simulate the behavior of robots without direct interaction with a physical robot. Under this category we include web robotics as a way of learning online using a web-based platforms for simulating robots, as e.g. in \cite{alvarez}. This latter is gaining momentum with offerings such as AWS RoboMaker\footnote{\url{https://aws.amazon.com/robomaker/}} which are cloud-based simulation services that enable robotics developers to run, scale, and automate simulation without managing any infrastructure. One of the most widely used simulators for ROS is Gazebo\footnote{\url{https://gazebosim.org/}} which provides 3D physics simulation for a large variety of sensors and robots. Gazebo is bundled into the full installation package of ROS, making it widely and easily available, and many robot manufacturers offer ROS packages specifically designed to support Gazebo. Other popular robotic simulators are Webots\footnote{\url{https://cyberbotics.com/}}, CoppeliaSim\footnote{\url{http://www.coppeliarobotics.com/}} and OpenRave\footnote{\url{http://www.osrobotics.org/osr/}}. Besides these, game engines are also being adapted to support robotic simulation such as, for instance, Unity\footnote{\url{https://github.com/Unity-Technologies/Unity-Robotics-Hub}}. Simulation based solutions are clearly useful and serve some important educational needs; however, the models on which they are based always have some limitations which can become apparent in a real world context. Further, adopting a simulation only approach does not give students experience with some of the more practical considerations associated with working with physical devices.

\textbf{Hardware-based learning} focuses on direct interaction and programming of physical robots. In some simple domains and for simple applications students can safely interact directly with the hardware without necessarily having first simulated the application behavior. One example of this is the LEGO® Robot Programming for kids program\footnote{\url{https://www.lego.com/en-gb/categories/coding-for-kids}} where kids build a robot, program it and interact with it; programming in this environment is based on a set of predefined tasks the robot can execute. Similar solutions based on compositions of predefined tasks resulting in more complex behaviours exist (e.g., the Blockly interface of the Niryo Ned robotic arm\footnote{\url{https://niryo.com/robotic-solution-education-research/}}). Such solutions, however, lack flexibility and the extensibility and customization capabilities required for real world robotics scenarios. To develop more realistic applications the use of programming languages such as Python, C++, MATLAB or ROS is a must. Moreover, in complex environments, where access to hardware is not always possible or too expensive, it becomes also mandatory to test the application behavior in a simulated environment first.


\textbf{Hybrid learning combining simulation and hardware-based learning} is a solution in which the robotic application can be  tested in a simulated environment and deployed on the physical devices in either a two-steps process or in hybrid manner. In the two-steps process where we keep simulations (first step) separate from testing on real hardware (second step). In doing so we have the advantages of less costs, reduced risks of damaging expensive hardware, reduced risks of damages to third persons and things. In a hybrid approach, concepts like  \textit{digital-twin} gain importance for developing robotic applications/tasks. A digital copy of a robotic  hardware can be used for visualization and control of the robot. In advanced solution, a digital-twin can be placed into a simulated environment while the actions and tasks are physically executed on the hardware. In this way, the simulated environment will provide inputs to the application in terms of environment (e.g., obstacles), sensing information (e.g., light, temperature), which allows to test applications in a close-to-real environment.

As the complexity of robotic applications is growing steadily, with the adoption of advanced analytic solutions such as  Artificial Intelligence, Semantic Navigation, Autonomous motion, new needs appeared in terms of computation, networking and storage resources. To cope with them, Cloud-based solutions started to see the light for computation offloading on the Cloud. The possibility for remotely controlling robotic systems further reduces costs for deployment, monitoring, diagnostic and orchestration of any robotic application. This, in turn, allows for building lightweight, low cost and smarter robots as the main computation and communication burden is brought to the cloud. Since 2010, when the Cloud Robotics term first appeared, a number of projects (e.g., RoboEarth \cite{roboearth} DAVinci \cite{davinci}) investigated the field pushing forward both research and products to appear on the market. Companies started investing in the field as they recognized the huge potential of cloud robotics. This lead to first open source cloud robotics frameworks appearing in recent years. An example of these is the solution from Rapyuta Robotics\footnote{\url{https://www.rapyuta-robotics.com/}}. Similarly, commercial solutions for developers have seen the light with the big players in the Cloud field joining the run (e.g., Amazon Robomaker and the Google Cloud Robotics Platform\footnote{\url{https://cloud.google.com/cloud-robotics/}}).




In our robotic application programming course, the objective is to teach students the use of ROS and application development addressing problems which typically arise in a robotics context, e.g. navigation and mapping, grasping of objects and perception. The students should be able to develop and perform some testing using only simulation environments before using their code on the physical robots. Further, embracing the Cloud Robotics paradigm, some components of the robotic application should run on the physical robots, while others should run on the cloud or the edge of the network. The objective of our system setup is that students can seamlessly transition their applications from the simulation environment to the real world context, while not having to address the troublesome issues associated with framework setup and networking which arise in such distributed systems. 

\section{Solution design and implementation}
\label{sec:solution_design}

\subsection{Possible Solutions}
\label{sec_3.1}

As the students required low-friction interaction with a simulation environment, a centralized hosted solution was required. Given that it had to run our specific labs and interact with networks specific to our environment, an off-the-shelf solution was not possible. Hence, we had different options for the hosted part of our solution:
\begin{itemize}
    \item \emph{Virtual Machines:} In this approach a dedicated VM is provided for each robot with ROS and an X-Windows session running in the VM Operating System;
    \item \emph{Containers running in dedicated Virtual Machines:} In this approach a dedicated VM is provided for each robot with ROS and an X-Windows session running in containers within the VM;
    \item \emph{Containers running with Kubernetes:} In this approach a container is created on the Kubernetes cluster for each robot - this container runs X-Windows and the necessary ROS processes.
\end{itemize}

The first approach was disregarded quickly as it is not sufficiently flexible for the educational context - students would have to install too many components and may need to perform non trivial troubleshooting in case of problems, typically with software dependencies. The second approach had the benefit that the solution could be developed, packaged into a container and tested a priori; the students could then easily install it on the provided VM using standard container management tools (Docker in this case). However, a limitation of this approach was that it required a dedicated GPU for each robot - while this was somehow manageable for one delivery of the course, with increased demand for GPUs within the organization, solutions which provided more efficient use of GPUs were preferred.

The Kubernetes based solution was attractive from the perspective of using modern, widely used container solutions, not needing to manage dedicated VMs and potentially making more efficient use of GPUs. Hence, we proceeded to realize a Kubernetes based solution.

\subsection{The ZHAW RAP Education Platform}
The ZHAW RAP Platform was built on the following technologies:
\begin{itemize}
    \item \emph{Openstack Cloud Infrastructure Platform:} this is our base platform on which supports management of compute and storage resources within our environment;
    \item \emph{K3S Container Platform:} this is a lightweight Kubernetes distribution which for us provided a good balance between ease of deployment and management, and modest resource utilization; 
    \item \emph{Cinder CSI Driver:} this is a specific storage driver which binds Kubernetes storage volumes to volumes in Openstack;
    \item \emph{Rancher:} the Rancher platform has multiple uses - we use it for user management and resource management on the Kubernetes cluster;
    \item \emph{Nvidia T4 GPUs:} We use server grade GPUs in our cluster.
\end{itemize}

The above are complex technologies with comprehensive documentation - it is not in the scope of this paper to include all aspects of bringing up a cluster comprising those technologies; rather the focus is on specific aspects which relate to providing the functionalities required to deliver the Robotics Educational content.

\subsubsection{Building containers}

Building a container which can run in this context is non trivial.  We used a base container which was designed for running GPU
backed X-Windows sessions on a Kubernetes cluster\footnote{https://github.com/ehfd/docker-nvidia-glx-desktop}. This base container
had already solved problems associated with running headless X-Windows on a GPU and providing a web interface to this using noVNC.

In our case, it was simply necessary to add Robotic specific packages, including ROS and Gazebo. This resulted in a working
container which students could use in their lab sessions. We developed a more sophisticated git based workflow in which we
create specific containers for each lab session, tailored to the focal point of the lab (e.g., running SLAM in a simulated environment,
investigating grasping mechanisms in simulation, investigating grasping with real robots, working with vision algorithms with real
device input etc).

The one specific consideration which had to be borne in mind related to nvidia driver versions: versions used in the container build process had to be consistent with those used on the Kubernetes cluster: this meant that we had to ensure that the VM on which we were
performing the container build had the same driver version as those installed on the cluser - also, we explicitly pinned the driver
versions on the Kubernetes nodes, ensuring that they were not upgraded automatically. This is some inflexible and means that GPU oriented 
containers will need to be rebuilt whenever they need to be deployed to a VM with newer nvidia driver versions. Going forward, it
means there will be friction associated with deploying GPU applications to nodes and it is likely we will need to label nodes in
the Kubernetes cluster with the nvidia driver versions and use this information accordingly when scheduling GPU workload to the cluster.

\subsubsection{Managing GPU resources}

As noted above, one of the key drivers for this approach is to support sharing of GPUs. Sharing nvidia GPUs in containerized environments
is evolving with the release of Multi-instance GPUs (MIG)\footnote{https://www.nvidia.com/en-us/technologies/multi-instance-gpu/} which is 
a promising solution which will support accurate control of GPU resources. Our approach, however, was to use a simpler solution based on 
technologies with which we already had experience.

The \texttt{nvidia-docker} runtime provides access to GPUs for containers running on a host -- any container running with this runtime will
have access to the GPU: it does not provide fine-grained control over these resources, however, meaning that any single container can 
consume all the resources of a single GPU.

We performed some rudimentary experimentation to determine how many sessions could share a single GPU: this comprised launching
different numbers of concurrent sessions, running the robotics tools and determining when the user experience was not sufficiently
responsive and the frame rate of the rendering started to drop. We found that it was possible to run 2 concurrent sessions on a
single GPU with appropriate performance for this context.

It was then necessary to devise a solution by which we could limit the amount of sessions active on a single node. The solution
involved specifying required CPU and memory resources for each pod such that no more than two could be scheduled concurrently on
any node - this was combined with use of the Kubernetes taint/tolerance mechanisms to ensure that only this workload could be
scheduled to these nodes. This did mean that there were specific VMs which could only be used with workloads which could tolerate
these taints, effectively limiting the use of these VMs and their GPUs to this educational activity. In future we may consider
other alternatives in which this workload could preempt other use of the GPUs such that they could be used when no educational
activities are taking place.

\subsubsection{Networking and communication considerations}

\begin{figure}[]
	\centering
    \includegraphics[scale=0.2]{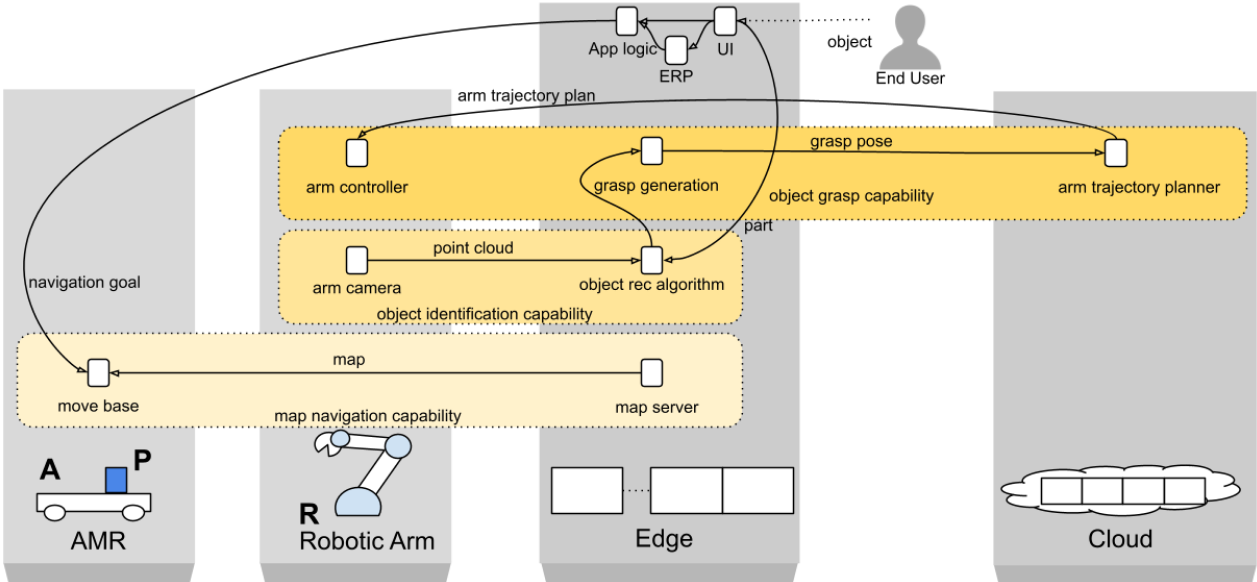}
	\caption{Distributed System for robotic applications.}
	\label{fig:dist}
\end{figure}


In our network configuration where the robotic hardware is on an internal network which is not externally accessible and the Kubernetes cluster is externally accessible, bidirectional communications could only be initiated by the devices. For ROS messages exchange between the hardware and the cloud/edge elements, a combination of a \texttt{rosbridge\_server}\footnote{\url{https://github.com/icclab/rosbridge_suite}} component on the cloud side and multiple \texttt{rosduct}\footnote{\url{https://github.com/icclab/rosduct}} instances running on the devices was used. For instance, in the pick and place setup the \texttt{rosduct} instances were running on a Raspberry Pi for controller messages, coordinate frames messages from the robotic arm, and the pointcloud and color images from the camera. The \texttt{rosduct} instances connect a websocket on the \texttt{rosbridge} having a Kubernetes Ingress as entrypoint.  In the navigation and mapping case, the \texttt{rosduct} instances were running directly on the Turtlebot 3 to exchange controller messages, messages from the scanner, coordinate frames and joint states messages with the \texttt{rosbridge\_server}.

The resulting system is sketched in Fig. \ref{fig:dist} with software components running on the hardware, and on the edge/cloud in the deployed container. For instance, for the navigation and mapping application, the \texttt{move\_base} component is running on the Turtlebot 3 Burger, whereas the \texttt{map\_server} ROS node, the visualization node and the application logic are running on the edge. Similarly, for the pick and place application, the controllers and \texttt{moveit} run on the Niryo arm and there is a Realsense D435 camera connected to a Raspberry Pi 3, which runs the application logic, the visualization component, object recognition and grasp generation logic.

\section{Demonstration and Evaluation}
\label{sec:demo_eval}
\begin{figure}[]
	\centering
    \subfigure[Gazebo simulator \label{fig:gazebo}]
    {\includegraphics[scale=0.12]{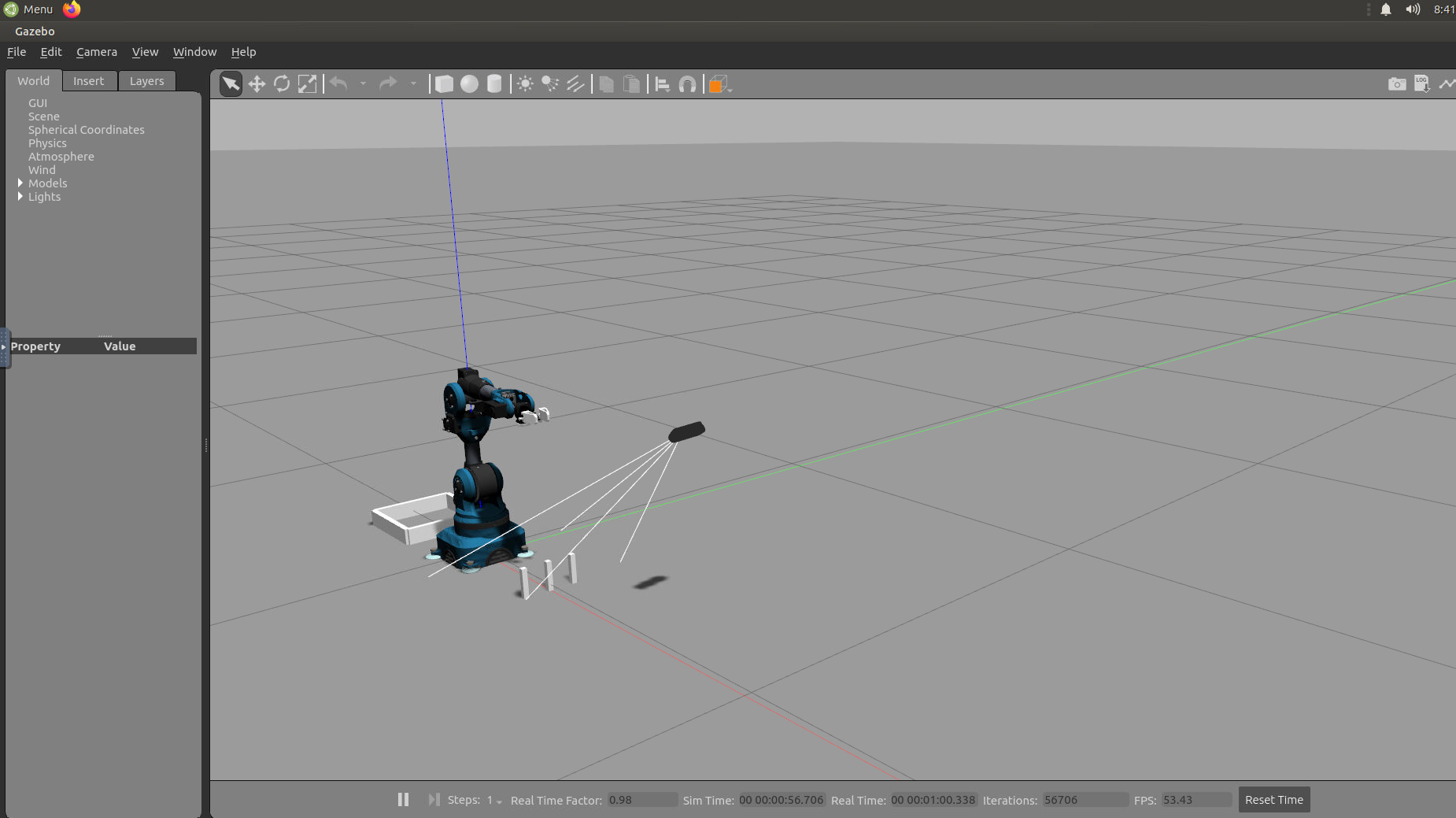}}
    \subfigure[RViZ visualization \label{fig:rviz}]
    {\includegraphics[scale=0.12]{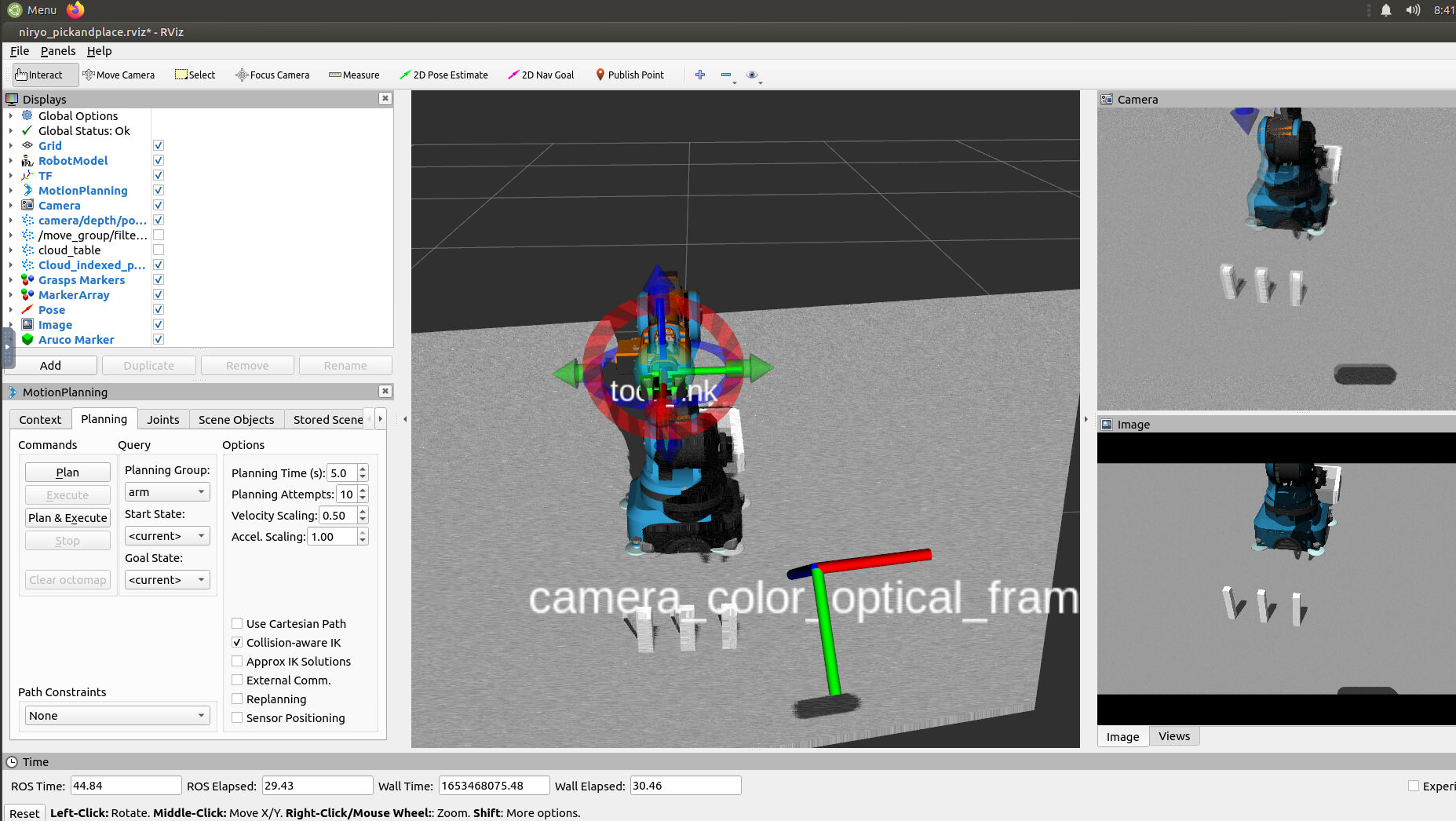}}
	\caption{Cloud-native simulation environment.}
	\label{fig:test1}
\end{figure}

\begin{figure}[]
	\centering
    \includegraphics[scale=0.3]{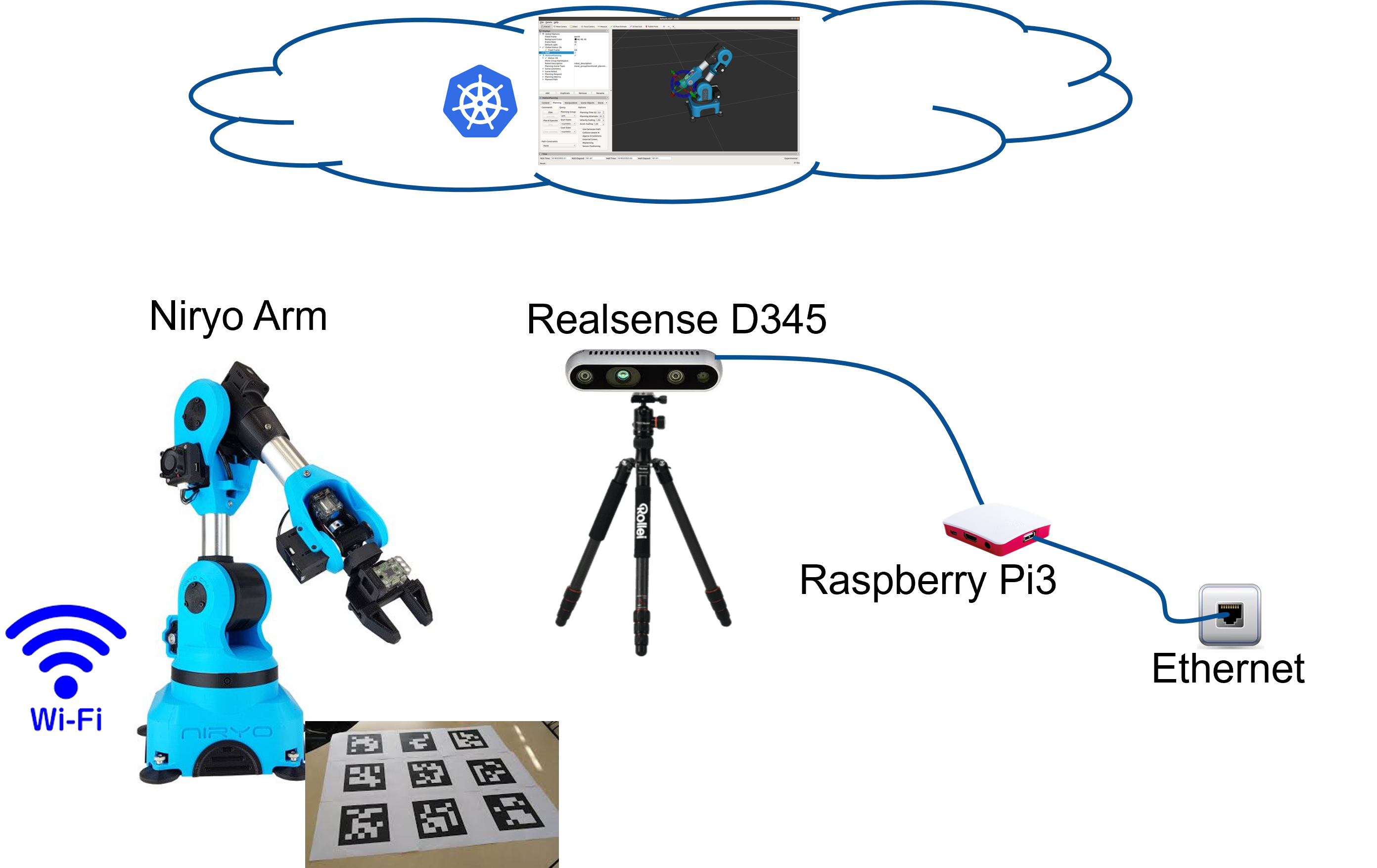}
	\caption{Distributed system setup for pick and place application.}
	\label{fig:test2}
\end{figure}

 Seven groups of three students on average shared the infrastructure we created to either run ROS applications in pure simulation (see Figs. \ref{fig:test1}) or in a mixed simulation and hardware setup. In particular, the hardware that was integrated for the mixed scenarios was the Niryo arm, a Raspberry Pi v3 with a Realsense D435 camera for the pick and place tasks and the Turtlebot 3 Burger for navigation and mapping of an unknown environment (see Fig. \ref{fig:test2}). 

In all of the labs, the students were easily able to deploy their lab environment on the Kubernetes cluster and start working on their tasks. To obtain some indicator of how well the system performed we measured the amount of Frames-Per-Second (FPS) which were obtained in the rendering - this gave a good estimate of how interactive the system was when all groups were working on the shared infrastructure. One component that heavily influences the system load is the Gazebo simulator. Frame rates of 34 - 60 FPS were observed in Gazebo, with the lower values being recorded during computationally heavy task executions. A second parameter we observed is the real time factor, which measures how fast the simulation time is running with respect to wall clock time. We observed that this parameter had values in the range of 0.80 - 0.98, with also in this case the lower end values observed in case of computational heavy tasks execution. This shows that the simulation environment operated slightly slower than the real world environment but the difference was small and hence not an issue in the lab environment.

\section{Open issues and challenges}
\label{sec:issues_challenges}

The solution described above meets many of the course requirements. However, some limitations
remain which mean it cannot be used in all possible scenarios; here we note the limitations
of the solution.

\subsubsection{Networking and Security}

In our configuration, the physical devices and the Kubernetes cluster were in
different network zones; further, as different policies applied to these zones, there 
were traffic restrictions between these network zones. This gave rise to some challenges 
in our context but, more generally, it is representative of many scenarios and as such 
it was important for us to find some solutions which can work within these constraints.

More specifically, the devices were on an internal network which was not externally
accessible; conversely, the Kubernetes cluster was externally accessible. In this
configuration, bidirectional communications could only be initiated by the devices.

Another limitation of our solution is that the Kubernetes cluster only supports
HTTPS connections on port 443; TCP connection to arbitary ports are not supported.
While this is supported in Kubernetes 
generally\footnote{https://kubernetes.github.io/ingress-nginx/user-guide/exposing-tcp-udp-services/}, 
our Kubernetes cluster is multi-purpose and hence there are limitations on introducing 
more sophisticated Kubernetes configurations, especially those which provide external 
connectivity. This specific limitation meant that communications using ROSTCP was not 
possible in our context.


\subsubsection{Container Images}

The container images which were built for this lab contained many substantial
components - X-Windows and a window manager, ROS, RViz, Gazebo and development
tools; as such, the resulting container images were large ($>$10GB). Working with 
such large container images does generate some friction - container build times
can be slow as can be pulling/pushing to/from remote registries. Slower container 
launch times could impact user experience; in our case, we ensured that each
Kubernetes worker node was pre-seeded with the appropriate lab container image
by simply running the container on all nodes. More sophisticated solutions
which support this exist such as Fledged\footnote{https://github.com/senthilrch/kube-fledged} 
or Dragonfly\footnote{https://d7y.io/} but we have not investigated these as yet.

\subsubsection{ROS Connectivity Solution}
As noted above, the current solution is based on a customized \texttt{rosbridge} websocket on Kubernetes pods and customized \texttt{rosduct} components on the robots. While this worked for the scope of our course, the solution can be further improved. We observed frequent unexpected disconnections from the websocket - the websocket was reestablished quickly so it was not unusable but it led to some performance degradation. Also, there were issues with some CBOR (Concise Binary Object Representation) encoding causing errors in the use of \texttt{rosduct} in combination with the \texttt{rosbridge} server - this requires further investigation as it can help reduce the amount of data transferred, ultimately making the system more responsive.

\section{Conclusion}
\label{sec:conclusion}

In this paper, we described our platform to support the teaching activities for Robotic Application Programming. The platform supports development of robotics applications and testing/validation in a simulation context before deploying to real world robots. All of this can be done within the same environment. We discussed the design choices made and the technologies we adopted to minimize the set up times and to support seamless ``simulation to real'' experience. The final setup was shown to provide good performance for the students while efficiently consuming shared GPU resources. Issues with the current solution which require further work have been identified and will be addressed for the next delivery of the module.

%


\begin{thebibliography}{1}

\bibitem{schina} Schina, D., Esteve-González, V., \& Usart, M. An overview of teacher training programs in educational robotics: characteristics, best practices, and recommendations. Educational Information Technologies, 2020. https://doi.org/10.1007/s10639-020-10377-z

\bibitem{alvarez}  Roldán-Álvarez, David, Sakshay Mahna, and José M. Cañas. "A ROS-based Open Web Platform for Intelligent Robotics Education." International Conference on Robotics in Education (RiE). Springer, Cham, 2021.

\bibitem{curto} Curto, Belén, and Vidal Moreno. "Robotics in education." Journal of Intelligent \& Robotic Systems 81.1 (2016): 3.

\bibitem{garcia} García, Sergio, et al. "Robotics software engineering: A perspective from the service robotics domain." Proceedings of the 28th ACM Joint Meeting on European Software Engineering Conference and Symposium on the Foundations of Software Engineering. 2020.

\bibitem{toff}
Toffetti, G., and Bohnert T. M. Cloud Robotics with ROS. In Robot Operating System (ROS), pp. 119-146. Springer, Cham, 2020.

\bibitem{koldberg} K. Goldberg and R. Siegwart, Eds., "Beyond webcams: an introduction to online robots." Cambridge, MA, USA: MIT Press, 2002.

\bibitem{inaba} M. Inaba, S. Kagami, F. Kanehiro, Y. Hoshino, and H. Inoue, "A platform for robotics research based on the remote-brained robot approach." I. J. Robotic Res., vol. 19, no. 10, pp. 933--954, 2000.

\bibitem{roboearth} M. Waibel, M. Beetz, J. Civera, R. D'Andrea, J. Elfring, D. Gálvez-López, K. Haussermann, R. Janssen, J. Montiel, A. Perzylo, B. Schiessle, M. Tenorth, O. Zweigle, and R. van de Molengraft, "Roboearth," Robotics Automation Mag., IEEE, vol. 18, no. 2, pp. 69--82, June 2011.

\bibitem{davinci} R. Arumugam, V. R. Enti, K. Baskaran, and A. S. Kumar, "DAvinCi: A cloud computing framework for service robots," in Proc. IEEE Int. Conf. Robotics and Automation. IEEE, May 2010, pp. 3084--3089.

\bibitem{karelekas} Karalekas, G., Vologiannidis, S., Kalomiros, J.: EUROPA–A ROS-based open platform for educational robotics. In: 2019 10th IEEE International Conference on Intelligent Data Acquisition and Advanced Computing Systems: Technology and Applications (IDAACS), vol. 1, pp. 452–457. IEEE, September 2019

\end{thebibliography}
%

\end{document}